\pdfoutput=1
\documentclass[11pt]{article}

\usepackage[preprint]{acl}

\usepackage{times}
\usepackage{latexsym}

\usepackage[T1]{fontenc}
\usepackage[utf8]{inputenc}

\usepackage{microtype}

\usepackage{inconsolata}

\usepackage{hyperref}       % hyperlinks
\usepackage{url}            % simple URL typesetting
\usepackage{booktabs}       % professional-quality tables
\usepackage{amsfonts}       % blackboard math symbols
\usepackage{nicefrac}       % compact symbols for 1/2, etc.
\usepackage{microtype}      % microtypography
\usepackage{xcolor}         % colors
\usepackage{times}
\usepackage{latexsym}
\usepackage{hyperref}
\usepackage{hwemoji}
\usepackage{color, colortbl}
\definecolor{Gray}{gray}{0.9}
\usepackage{array}
\usepackage{booktabs}
\usepackage{multirow}
\usepackage{tcolorbox}
\usepackage{amsmath}
\usepackage{amssymb}
\usepackage{graphicx}
\usepackage{enumerate}
\usepackage{diagbox}
\usepackage{mathtools}
\usepackage{csquotes}
\usepackage{paralist}
\usepackage{tabularx}
\usepackage{lipsum}
\usepackage{ragged2e}
\usepackage{wrapfig}

\usepackage{inconsolata}
\newcommand{\name}[1]{\textsc{EcomEdit}}

\newcommand{\dataset}{\textsc{EcomEdit}}

\title{\textsc{EcomEdit}: An Automated E-commerce Knowledge Editing Framework for Enhanced Product and Purchase Intention Understanding}

\author{Ching Ming Samuel Lau\thanks{Equal Contribution},
Weiqi Wang${^*}$,
Haochen Shi,\\
\textbf{Baixuan Xu,
Jiaxin Bai,
Yangqiu Song}\\
Department of Computer Science and Engineering, HKUST, Hong Kong SAR, China\\
\texttt{cmslau@connect.ust.hk, \{wwangbw, yqsong\}@cse.ust.hk}\\
}

\begin{document}
\maketitle

\begin{abstract}
Knowledge Editing (KE) aims to correct and update factual information in Large Language Models (LLMs) to ensure accuracy and relevance without computationally expensive fine-tuning. 
Though it has been proven effective in several domains, limited work has focused on its application within the e-commerce sector.
However, there are naturally occurring scenarios that make KE necessary in this domain, such as the timely updating of product features and trending purchase intentions by customers, which necessitate further exploration. 
In this paper, we pioneer the application of KE in the e-commerce domain by presenting \name{}, an automated e-commerce knowledge editing framework tailored for e-commerce-related knowledge and tasks. 
Our framework leverages more powerful LLMs as judges to enable automatic knowledge conflict detection and incorporates conceptualization to enhance the semantic coverage of the knowledge to be edited. 
Through extensive experiments, we demonstrate the effectiveness of \name{} in improving LLMs' understanding of product descriptions and purchase intentions. 
We also show that LLMs, after our editing, can achieve stronger performance on downstream e-commerce tasks. 
Our code, data, and models will be released upon acceptance.
\end{abstract}

\section{Introduction}
Knowledge Editing (KE) aims to provide LLMs with timely and accurate updates to ensure the factual accuracy and comprehensiveness of their knowledge~\cite{yao2023editinglargelanguagemodels,DBLP:journals/corr/abs-2310-16218}. 
This need arises because fine-tuning is computationally expensive, particularly as the number of parameters increases. 
Additionally, fine-tuning can lead to overfitting when based on a limited number of samples, which restricts the model's ability to generalize to new information~\cite{wang2024SafeEdit,DBLP:conf/acl/JuCY0DZL24,DBLP:conf/acl/Wang000XZL24}.

\begin{figure}[t]
    \centering
    \includegraphics[width=1\linewidth]{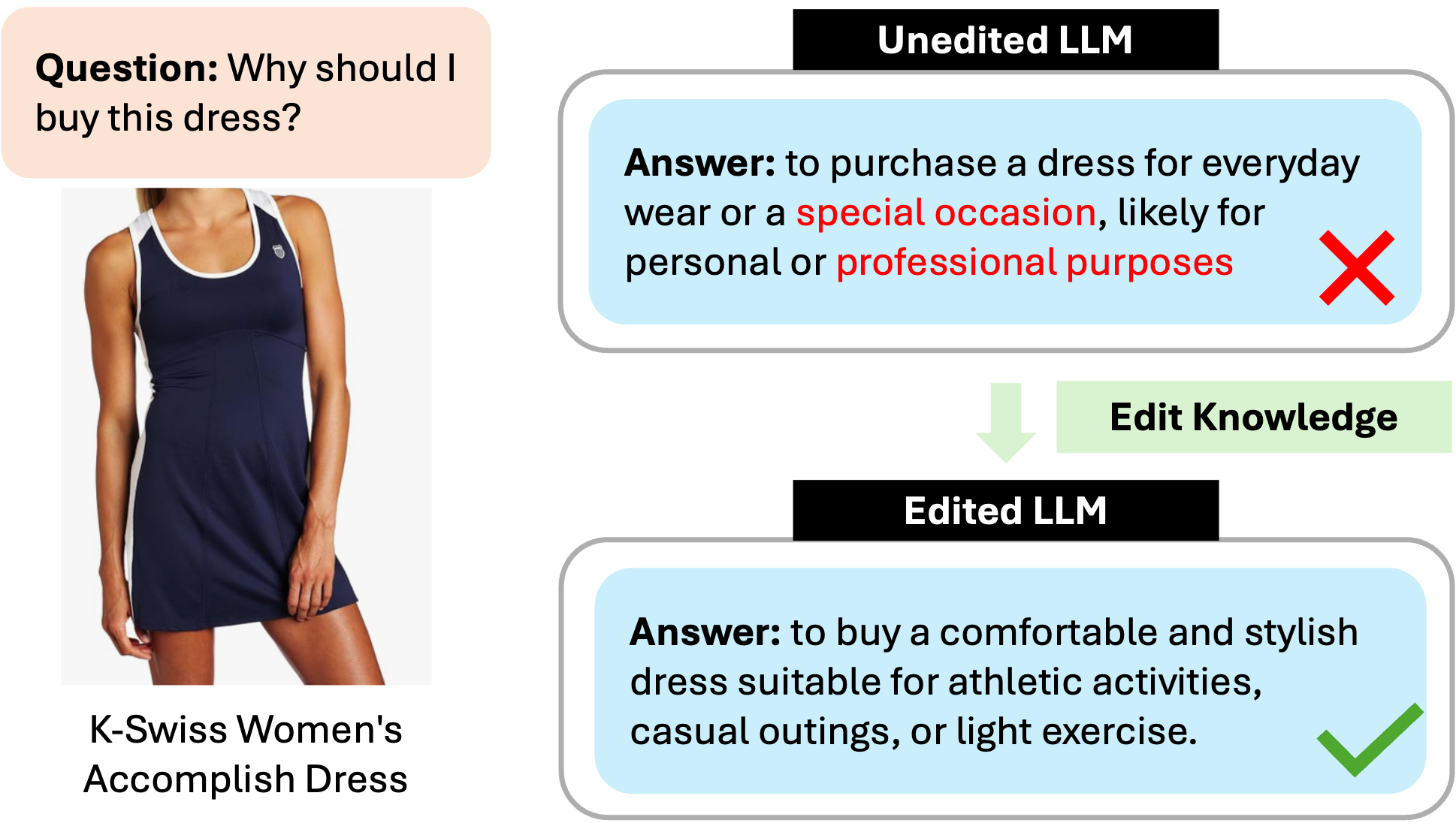}
    \vspace{-0.1in}
    \caption{An example of editing purchase intention to make the LLM more of an e-commerce expert.}
    \vspace{-0.2in}
    \label{fig:intro_example}
\end{figure}

However, existing literature primarily focuses on applying knowledge editing to straightforward factual knowledge, such as knowledge found in Wikipedia~\cite{zhang2024comprehensive, wang2023easyedit}. 
This presents a significant gap in the application of KE within the e-commerce domain, despite its considerable potential. 
As LLMs become increasingly integrated into e-commerce as intelligent shopping assistants~\cite{palenmichel2024investigatingllmapplicationsecommerce,DBLP:conf/eacl/ChanCWJFLS24}, the reliance on these models for accurate and relevant information intensifies, making it essential for them to remain up-to-date.
Nevertheless, LLMs often face implicit challenges such as outdated product descriptions, inaccurate feature representations, and difficulties in understanding customer intent. 
For instance, as illustrated in Figure~\ref{fig:intro_example}, an LLM might misinterpret a product's purpose, leading to the dissemination of incorrect information to users. 
While knowledge can be injected through direct fine-tuning, this approach is typically cost-prohibitive when products are frequently updated or when new products are introduced.

To tackle these issues and explore the feasibility of applying KE methods in the e-commerce domain, we propose~\name{}, an automated knowledge editing framework that leverages a powerful LLM as a judge to identify and analyze the weaknesses of smaller e-commerce expert models and supervise the editing process.
By applying our framework to real-world e-commerce data, we construct several knowledge editing datasets tailored to specific models, with minimal human intervention. 
We then experiment with a selection of representative knowledge editing methods to validate our framework by presenting (1) expert annotation results that demonstrate improved accuracy in product and intention understanding in LLMs and (2) enhanced performance on e-commerce intention understanding tasks.
To the best of our knowledge, we are the first to explore knowledge editing methods in the e-commerce domain, providing valuable insights into applying such techniques in a field where knowledge updates frequently.
We will make our data and models publicly available upon acceptance.

\section{Related Works}

\noindent\textbf{Knowledge Editing.}
The need for knowledge editing arises from various scenarios, such as correcting factual errors, fixing model biases, detoxifying outputs, and updating knowledge in response to new information. 
\citet{mitchell2022memorybasedmodeleditingscale} introduced SERAC, a memory-based method for scalable model editing that maintains overall performance. 
Similarly, \citet{sinitsin2020editableneuralnetworks} utilized meta-learning to enhance editing efficiency. 
\citet{meng2023locatingeditingfactualassociations} proposed ROME to directly adjust parameters for editing factual associations within GPT models.
Meanwhile, several benchmarks have been constructed~\cite{wang2024SafeEdit, wang2024editingconceptualknowledgelarge, cheng2024editmultimodallargelanguage} to evaluate the impact of various knowledge editing techniques on model robustness and generalization.
In this work, we step forward by bringing KE into e-commerce.

\noindent\textbf{Product and Intention Understanding.}
Previous works have demonstrated the effectiveness of LLMs in understanding and describing product features, as well as uncovering user intentions through product classification, subgraph reasoning, and multimodal approaches that integrate visual and textual information~\cite{gholamian2024llmbasedrobustproductclassification, shi2024llmpoweredexplanationsunravelingrecommendations, ROUMELIOTIS2024100056, jiang-etal-2024-hallucination, xu2024mindmultimodalshoppingintention,DBLP:conf/acl/YuWLBSLG0Y23,DBLP:conf/sigmod/YuLML0GSGZYL24,EcomScript,DBLP:journals/corr/abs-2402-18169}. 
These works highlight the potential of LLMs in e-commerce applications. 
To further strengthen LLM's potential in e-commerce, in this work, we pioneer knowledge editing within the e-commerce domain for enhanced product and intention understanding.

\section{\name{} Framework}
\begin{figure*}[h]
    \centering
    \includegraphics[width=0.65\linewidth]{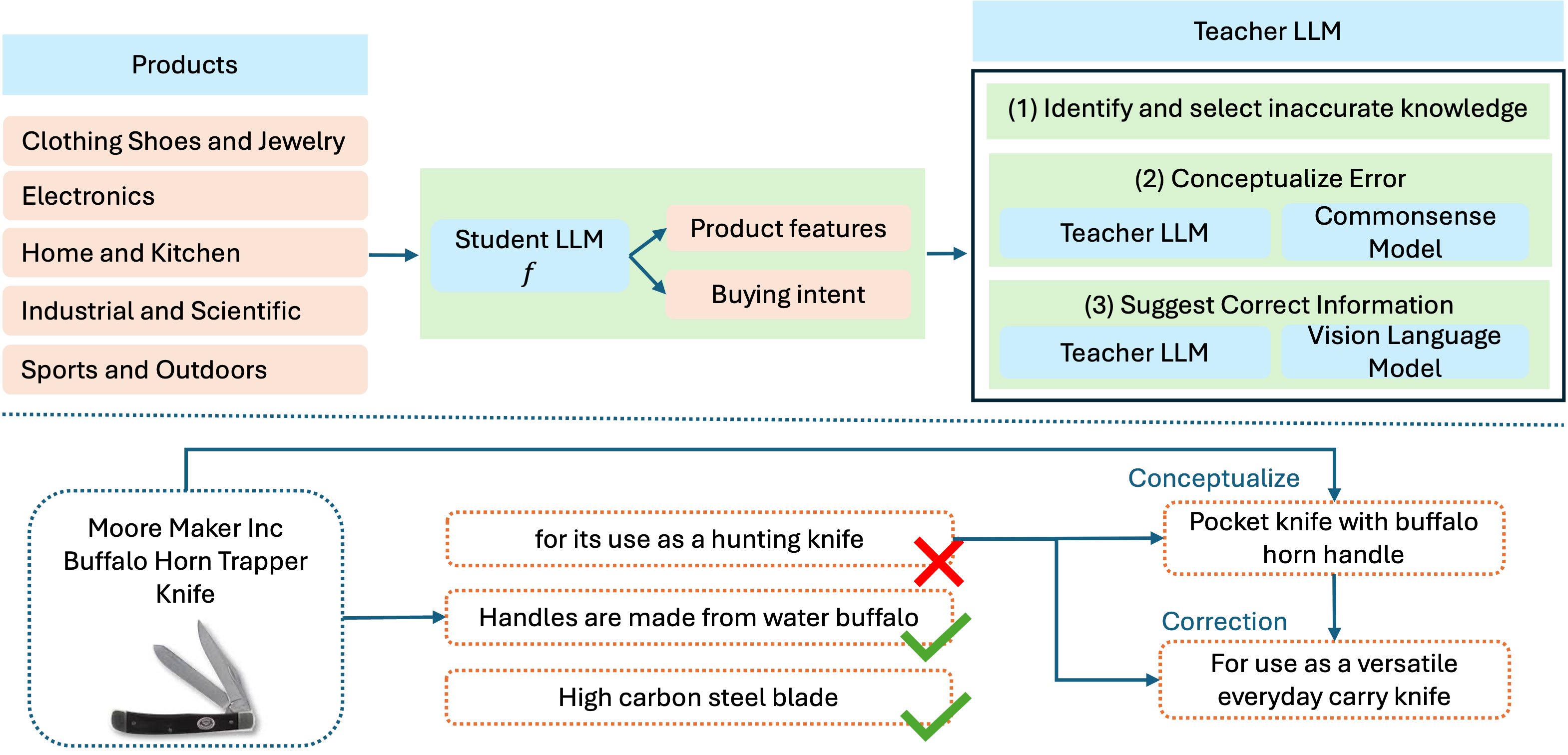}
    \vspace{-0.10in}
    \caption{An overview of our proposed KE pipeline (\name{}).}
    \vspace{-0.2in}
    \label{fig:flow}
\end{figure*}

In this section, we introduce our proposed framework, as shown in Figure~\ref{fig:flow}.
To ensure the practicality of our method, we randomly selected 2,000 products from five representative categories in the Amazon Review Dataset~\cite{hou2024bridging} as the raw data for verifying our framework.

\subsection{Feature and Intention Examination}
To examine the weaknesses and erroneous knowledge within current LLMs, we select four representative open-source LLMs: LLaMa3-8B~\cite{llama3modelcard}, Mistral-7B~\cite{jiang2023mistral7b}, Falcon-7B~\cite{almazrouei2023falconseriesopenlanguage}, and Gemma-7B~\cite{gemmateam2024gemmaopenmodelsbased}. 
We prompt them to generate product features and purchase intentions for each product. 
This step helps us measure the models' understanding of the items and their ability to propose relevant and accurate attributes and purchase motivations.

\subsection{Supervising KE by a Judge LLM}
We then enlist help from a more powerful LLM, GPT-4o~\cite{GPT4o}, as a judge model, to automate the KE process.
Specifically, we ask it to distinguish the correctness of the previously collected outputs regarding product features and purchase intentions. We then ask the judge LLM to check (1) whether the proposed feature is sensible and relevant to the item, and (2) whether the intention to buy aligns with the product and feature.

We filter out the cases where the judge LLM identifies errors in the proposed features or buying intentions. These mistakes serve as candidates for model editing, as they represent situations where the smaller LLMs have failed to provide accurate or coherent information. For each product that contains these incorrect features or intentions, we engage in a conceptualization process~\cite{DBLP:conf/acl/WangFXBSC23,DBLP:conf/emnlp/WangF0XLSB23,DBLP:conf/acl/0001FLS0XWBLJCS24,DBLP:journals/corr/abs-2406-10885,DBLP:journals/corr/abs-2406-02106}. Specifically, we ask the judge LLM to generate concepts to describe the product, aiming to include more types and categories of the product, which may also have incorrect descriptions and purposes~\cite{DBLP:journals/ai/HeFWS24}. For example, Winsome Wood Assembled Set of 2 Kids Chairs, White will be conceptualized as White Wood Chairs and Kids Wood Assembled Set of 2 Chairs, given that they all incorrectly feature Adjustable Height.

The generated concepts are then evaluated for their plausibility by aligning them with common sense and real-world knowledge. We verify whether the product’s features and buying intentions logically relate to its concept and overall functionality using VERA, a general-purpose plausibility estimation model~\cite{liu2023verageneralpurposeplausibilityestimation}. If the association is plausible, it is kept as a valid reference. If the generated product-feature-intention associations do not make sense in a commonsense context, or if they are inconsistent with the product’s core attributes, we mark them for model editing. Additionally, we aim to edit the incorrect product concepts themselves, refining them to be more accurate and aligned with the product’s attributes.

\subsection{Editing Features and Intentions}
Finally, to update the incorrect features and intentions, we utilize LLaVa \cite{liu2023improvedllava} and GPT-4o, as they have a more comprehensive understanding of the products in terms of textual descriptions and visual information. We prompt them to generate the correct product features and intentions, thereby providing ground truth data for KE.

Meanwhile, we define locality and portability evaluations for our editing task. Locality ensures that the edits are localized to the erroneous feature/intention pairs without negatively impacting unrelated knowledge. We establish locality through Distracting Neighbor \cite{yao2023editinglargelanguagemodels} by selecting product dimensions and descriptions that should remain unaffected after editing. Portability evaluates whether the edits generalize to similar cases, ensuring that the corrected knowledge is applicable beyond specific instances. To assess this, we use subject replacement, as done by~\citet{yao2023editinglargelanguagemodels}, to evaluate portability through synonym replacement by prompting GPT-4o.

\section{Experiments and Analysis}
We first present statistics of our collected data through~\name{}.
In total, we collect 18,925 samples for E-Commerce KE, in which 14,712 of them are features, and 4,213 of them are for intentions editing. 
Each sample is linked to a student LLM output that contains an incorrect feature or intention.
Statistics are shown in Table~\ref{table:data_stat}.
We have also constructed 2,298 locality samples and 2,782 portability samples for evaluation.
% Furthermore, products within the dataset are categorized into distinct types, including Sports and Outdoors (5,435 products), Industrial and Scientific (4,059 products), Electronics (3,500 products), Clothing, Shoes, and Jewelry (3,115 products), and Home and Kitchen (2,816 products). 
% Details are shown in Table \ref{table:data_stat}. 

\begin{table}[t]
\centering
\small
\resizebox{0.8\linewidth}{!}{%
\begin{tabular}{r r r | r}
\toprule
\textbf{Product Category} & \textbf{Feature} & \textbf{Intention} & \textbf{ Total} \\
\midrule
\textbf{Clothing Shoes and Jewelry} & 2,397 & 718 & 3,115 \\
\textbf{Electronics} & 2,888 & 612 & 3,500 \\
\textbf{Home and Kitchen} & 2,131 & 685 & 2,816 \\
\textbf{Industrial and Scientific} & 2,895 & 1,164 & 4,059 \\
\textbf{Sports and Outdoors} & 3,895 & 1,540 & 5,435 \\
\midrule

\textbf{Total} & 14,206 & 4,719 & 18,925 \\
\bottomrule
\end{tabular}%
}
\caption{\dataset{} dataset statistics.}
\vspace{-0.2in}
\label{table:data_stat}
\end{table}

\subsection{Intrinsic Evaluation}
In our human evaluation, we conducted a comprehensive assessment on \dataset{}. We systematically removed unreasonable or irrelevant data to ensure high-quality input. A sample of 1,000 entries from \dataset{} was evaluated whether the product has the corrected feature/intention mapping, achieving an accuracy of 90.3\%. 

\begin{table*}[htb]
\small
\renewcommand{\arraystretch}{0.8} % Increase row height
\setlength{\tabcolsep}{3pt} % Increase column padding
\centering
\resizebox{0.8\linewidth}{!}{
\begin{tabular}{c c c c c c c | c c}
\toprule
\multicolumn{2}{c}{\textbf{Methods}} & \textbf{Llama-3-8B} & \textbf{Mistral-7B} & \textbf{Falcon-7B} & \textbf{Gemma-1.1-7B} & \textbf{Total/Avg.} & \textbf{Feature} & \textbf{Intention} \\ 
\midrule
\multicolumn{2}{c}{\textbf{\#}} & \textbf{2,134} & \textbf{2,660} & \textbf{7,211} & \textbf{6,920} & \textbf{18,925} & \textbf{14,206} & \textbf{4,719} \\ 
\midrule
\multicolumn{1}{c}{\multirow{3}{*}{\textbf{FT}}} & \textbf{REL} & 57.79 & 60.02 & 55.59 & 27.82 & \textbf{46.31} & \textbf{46.84} & 45.10 \\
\multicolumn{1}{c}{} & \textbf{LOC} & 40.68 & 95.84 & 11.98 & 98.62 & \textbf{58.68} & 57.22 & \textbf{61.31} \\
\multicolumn{1}{c}{} & \textbf{POR} & 56.21 & 55.25 & 50.58 & 14.53 & \textbf{38.69} & 38.36 & \textbf{40.21} \\ 
\midrule
\multicolumn{1}{c}{\multirow{3}{*}{\textbf{LORA}}} & \textbf{REL} & 62.04 & 64.58  & 30.84 & 36.62  & \textbf{41.21} & \textbf{37.18} & 25.30 \\
\multicolumn{1}{c}{} & \textbf{LOC} & 1.93 & 2.22 & 1.67 & 1.96 & \textbf{1.88} & 1.83 & \textbf{2.19} \\
\multicolumn{1}{c}{} & \textbf{POR} & 13.57 & 8.05 & 14.38 & 22.13 & \textbf{16.23} & \textbf{18.21} & 10.29 \\ 
\midrule
\multicolumn{1}{c}{\multirow{3}{*}{\textbf{MEMIT}}} & \textbf{REL} & 77.06 & 90.11 & 97.60 & 52.15 & \textbf{77.61} & \textbf{83.76} & 84.93 \\
\multicolumn{1}{c}{} & \textbf{LOC} & 89.72 & 82.09 & 72.46 & 92.06 & \textbf{82.93} & 78.93 & \textbf{83.09} \\
\multicolumn{1}{c}{} & \textbf{POR} & 56.54 & 70.19 & 63.70 & 18.22 & \textbf{47.17} & \textbf{56.33} & 53.62 \\ 
\midrule
\multicolumn{1}{c}{\multirow{3}{*}{\textbf{ROME}}} & \textbf{REL} & 95.44 & 87.66 & 98.42 & 89.46  & \textbf{93.30} & 94.38 & \textbf{94.50} \\
\multicolumn{1}{c}{} & \textbf{LOC} & 72.17 & 81.02 & 77.09 & 79.28 & \textbf{77.89} & 77.21 & \textbf{78.32} \\
\multicolumn{1}{c}{} & \textbf{POR} & 73.69 & 69.15 & 61.55 & 64.84 & \textbf{65.19} & \textbf{56.33} & 53.62 \\
\bottomrule
\end{tabular}%
}
\vspace{-0.1in}
\caption{Experiment Results with Feature Intention Comparsion}
\label{table:combined_experiment_result}
\vspace{-0.2in}
\end{table*}

\subsection{Extrinsic Evaluation}
We ran experiments on \dataset{} by editing the respective 4 LMs through existing Kowledge Editing methods, namely fine-tuning (FT), Low-Rank Adapation (LoRA), Rank-One Model Editing (ROME), and Mass-Editing Memory in a Transformer (MEMIT).
Table \ref{table:combined_experiment_result} shows a comparative analysis of various methods applied to different language models, including Llama-3-8B, Mistral-7B, Falcon-7B, and Gemma-1.1-7B. Each method is evaluated based on three key metrics: relevance (\textbf{REL}), localization (\textbf{LOC}), and portability (\textbf{POR}). 
Notably, \textit{MEMIT} and \textit{ROME} consistently demonstrates superior performance across most metrics, particularly in relevance and localization, indicating its effectiveness in enhancing model capabilities. In contrast, \textit{FT} and \textit{LoRA}  shows comparatively lower scores, suggesting a need for further optimization. It also highlights that \textit{LoRA} suffers from overfitting in order to edit samples successfully. 
% This aligns with findings from other studies \textcolor{blue}{[add citation]}, which also report significant improvements in model adaptability and contextual accuracy when using knowledge editing methods.
\textit{Mistral-7B} is the easiest to edit among all LLMs, exhibiting the highest average reliability while effectively maintaining locality and portability. In contrast, \textit{Gemma-1.1-7B} is the most difficult to edit, struggling with reliability and often failing to preserve context in various scenarios. This limitation can be attributed to its underlying architecture, which may not be as optimized for editing as its counterparts. Both \textit{Llama-3-8B} and \textit{Falcon-7B} also present challenges in editing, particularly in maintaining locality and portability across different contexts. Editing knowledge in the context of e-commerce remains a significant challenge as the effectiveness of existing knowledge editing methods varies widely across different model architectures.

\subsection{Downstream Performance}

\begin{figure}[t]
    \centering
    \includegraphics[width=0.8\linewidth]{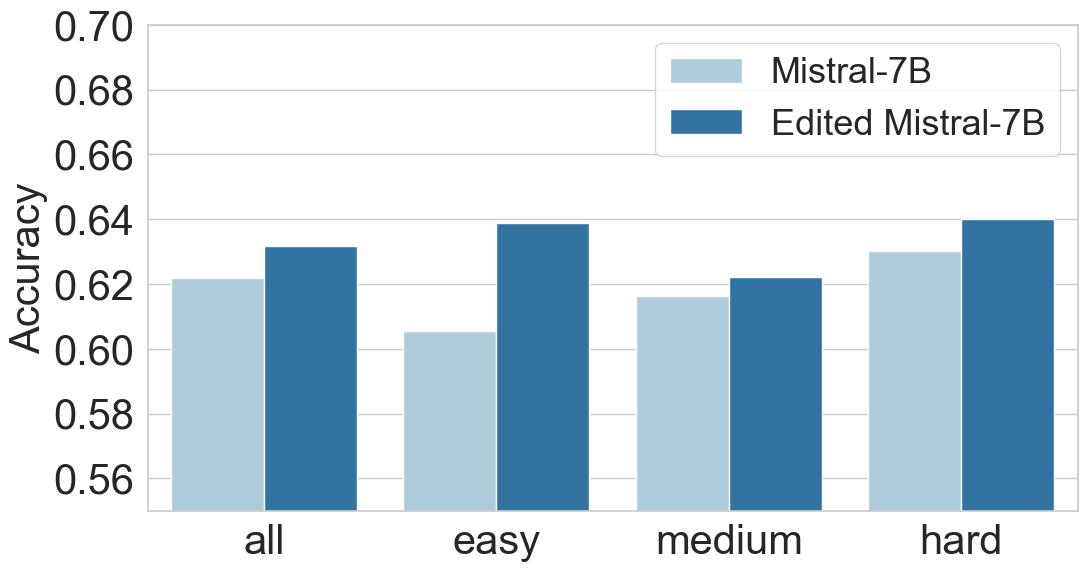}
    \vspace{-0.1in}
    \caption{\name{}-Mistral's performance on IntentionQA task.}
    \vspace{-0.2in}
    \label{fig:downstream}
\end{figure}

To evaluate the capabilities of \dataset{}, we tested the edited model using IntentionQA \cite{ding2024intentionqabenchmarkevaluatingpurchase}. Figure \ref{fig:downstream} shows a comparison of \textit{Mistral-7B} performance before and after 1,000 samples of edit using \textit{MEMIT} on Task 1. The results demonstrate a significant improvement, indicating that the edited model has enhanced its ability to understand and respond to user intentions effectively. This improvement shows the effectiveness of \dataset{} as well as the knowledge editing strategies.

\subsection{Comparing Feature and Intention Editing}
From Table \ref{table:combined_experiment_result}, it’s easier to edit features than intentions, as the reliability for features are typically higher across models.
For example, \textit{FT} shows a reliability score of 46.84 for features compared to 45.10 for intentions. 
Although the difference is minimal, it indicates a slight advantage in making edits to features. 
However, the locality for intention is greater than that for features. \textit{MEMIT} has a locality score of 83.09 for intentions but only 78.93 for features. 
This suggests that modifications to intentions are less likely to inadvertently affect unrelated components, allowing for more precise adjustments. 
Consequently, editing intentions can be more targeted and specific, leading to a more focused impact on related outcomes.
Regarding portability, it is also higher for features than for intentions, as seen in \textit{LoRA}, \textit{ROME} and \textit{MEMIT}. 
This means that while intentions may be less reliable, the edits made to features can be more easily generalized to similar cases, ensuring that the corrected knowledge applies effectively across various contexts. 
This nuanced understanding of feature to intention editing highlights the trade-offs between ease of modification and specificity in impact.

\section{Conclusions}
In this work, we addressed the challenge of knowledge editing within the E-commerce domain by developing \dataset{}, a specialized dataset designed to enhance the accuracy and relevance of product features and customer intentions. Our research highlighted the limitations of existing large language models in maintaining up-to-date product information and understanding user intent, which can significantly impact the shopping experience.

\clearpage
\section*{Limitations}
\name{} has three main limitations.
First, it contains only around 16,000 products across five categories, which restricts the generalizability of our findings to a broader range of products and categories in the e-commerce domain. 
Additionally, the edited knowledge relies on the product understanding of the judge LLM. While our human evaluations have demonstrated its accuracy, this performance may differ from our human understanding. 
Finally, our experiments on current knowledge editing methods are limited. 
In our future work, we would include more products and categories into constructing the benchmark and experiment more knowledge editing methods to evaluate \dataset{}.

\section*{Ethics Statement}
The dataset curation process for our project involves careful measures to eliminate offensive content in both prompts and generated responses from large language models (LLMs). We clearly instruct the LLMs to avoid generating material that violates personal privacy, promotes violence, hate speech, or self-harm~\cite{DBLP:journals/corr/abs-2406-11149,DBLP:conf/emnlp/ChengQCFWCRGZSZ23,DBLP:conf/nips/BaiLW0S23}. A manual review of 500 random data entries showed no offensive content, leading us to conclude the dataset is safe and poses no societal risks.
All resources are publicly available and will be shared under the MIT license for research purposes. The Amazon Review Dataset used is under a CC-SA license, allowing its research use, and we access open-source models through the Hugging Face Hub, complying with all licensing terms. Expert annotators contribute voluntarily to the research.

\bibliography{custom}

\appendix
\label{sec:appendix}

\section{Product Category Selection to construct \dataset{}}
We selected 5 product categories that represent a diverse range of items commonly encountered in E-commerce. These categories were chosen to balance general consumer interest and industry-specific products, ensuring that the dataset captures different purchasing intentions.
\begin{table}[hbt]
\small
\centering
\resizebox{0.5\linewidth}{!}{%

\begin{tabular}{l}
\toprule
 Clothing Shoes and Jewelry
\\
Electronics
 \\
Home and Kitchen
 \\
Industrial and Scientific
 \\
 Sports and Outdoors \\
 \bottomrule
\end{tabular}%
}
\end{table}

\section{Student LLM Prompt Template}
In Section 3.2, student LLMs will generate prod
uct features and intentions to buy for each of the
selected items. We used to following prompt:

\begin{table}[hbt]
\centering
\small
\resizebox{1\linewidth}{!}{%
\begin{tabular}{p{0.2\linewidth} p{0.8\linewidth}}
\toprule
\textbf{Intention} & \textit{A customer buys a product: \texttt{\{name\}}. What is the intention of buying it? \newline 
Please be concise and ONLY answer in ONE sentence. Start with 'The intention of buying this is to'.} \\
\midrule
\textbf{Feature} & \textit{Consider this product: \texttt{\{name\}}. \newline 
What are the features of it? \newline 
Please ONLY give AT MOST 3 features and start each feature with a new line.}  \\ 
\bottomrule
\end{tabular}%
}
\end{table}

\section{Judge LLM Prompt Template}
\subsection{Answer Checking}
In Section 3.2, Judge LLM will check whether th product features and intentions is correct. We used to following prompt:

\begin{table}[hbt]
\centering
\small
\resizebox{1\linewidth}{!}{%
\begin{tabular}{p{0.2\linewidth} p{0.8\linewidth}}
\toprule
\textbf{Intention} & \textit{A customer buys a product: \texttt{\{name\}}. Consider this product: \texttt{\{name\}}. \newline 
Do you think the product's \texttt{\{detail\_key\}} is \texttt{\{detail\_value\}}? \newline 
Please first answer yes or no. If it is yes, just return 'yes'. If it is no, please provide a brief explanation and corrected product detail. \newline 
Answer:} \\
\midrule
\textbf{Feature} & \textit{Consider this product: \texttt{\{name\}}. \newline 
Do you think it has this feature: \texttt{\{feature\}}? \newline 
Please first answer yes or no. If it is yes, just return 'yes'. If it is no, please provide a brief explanation and corrected features. \newline 
Answer:}  \\ 
\bottomrule
\end{tabular}%
}
\end{table}

\subsection{Propose Correct Features/Intention}
To propose the correct features and intention for Section 3.7, we use the following prompt:
\begin{table}[hbt]
\centering
\small
\resizebox{1\linewidth}{!}{%
\begin{tabular}{p{0.2\linewidth} p{0.8\linewidth}}
\toprule
\textbf{Intention} & \textit{Consider this product: \newline 
\texttt{\{name\}} \newline 
Currently, somebody has identified a wrong intention for buying this item: \texttt{\{feature\_or\_intention\}} \newline 
Please suggest a better, modified, and correct intention.} \\
\midrule
\textbf{Feature} & \textit{Consider this product: \newline 
\texttt{\{name\}} \newline 
Currently, somebody has identified a wrong feature: \texttt{\{feature\_or\_intention\}} \newline 
Please suggest a better, modified, concise, and most importantly, a correct feature.}  \\ 
\bottomrule
\end{tabular}%
}
\end{table}

\section{Conceptualize Prompt Template}
To conceptualize the wrong product-feature or product-intention pair, we use the following prompt:

\begin{table}[hbt]
\centering
\small
\resizebox{1\linewidth}{!}{%
\begin{tabular}{p{0.8\linewidth}}
\toprule
\textit{Please replace this product with another term: \texttt{\{product\}}. \newline 
You should not change the meaning of it. You can use a synonym or a general term. \newline 
It should be the same as the original product, which all of them should not have this feature/intention: \texttt{\{feature\_or\_intention\}}. \newline 
Please return at most 5 conceptualized products/product categories. They should be reasonable, and each of them should be separated by a new line.} \\
\bottomrule
\end{tabular}%
}
\end{table}

\section{Locality and Portability Prompt Template}
For locality and portability examples generation, we use the following prompt on \textit{Subject Replace} and \textit{Distracting Neighbor}:
\begin{table}[hbt]
\centering
\small
\resizebox{1\linewidth}{!}{%
\begin{tabular}{p{0.2\linewidth} p{0.8\linewidth}}
\toprule
\textbf{Portability: Subject Replace} & \textit{Please replace this subject with another term: \texttt{\{product\}}.\newline You should not change the meaning of it. You can use a synonym or a general term. Please only return the new subject.} \\
\midrule
\textbf{Locality: Distracting Neighbor} & \textit{Consider this product: \texttt{\{product\}}.\newline Here is a description of the product: \texttt{\{description\}}. \newline Please construct a sentence using the description based on the following template: The [ATTRIBUTE] of [PRODUCT] is xxx. \newline Please make sure the attribute is easily inferable from the product name.}  \\ 
\bottomrule
\end{tabular}%
}
\end{table}

\section{Benchmark Comparisons}
Several benchmark has been developed previously, yet their effectiveness and applicability can vary significantly across different tasks and domains. We have compared \dataset{} with the following benchmarks:

\begin{table}[hbt]
\centering
\small
\resizebox{1\linewidth}{!}{%
\begin{tabular}{r r r r}
\toprule
\textbf{Benchmark} & \textbf{\#} & \textbf{Domain} & \textbf{Construction} \\
\midrule
\textbf{Counterfact} & 21919 & General & Factual Replacement from Wiki \\
\textbf{MQuAKE} & 9218 & General & Counterfact with multi-hop \\
\textbf{ZsRE} & 244,173 & General QA & Relationship Extraction \\
\textbf{EComEdit} & 18,925 & E-Commerce & Identify LLM Mistakes \\ 
\bottomrule
\end{tabular}%
}
\caption{Benchmark Comparsions}
\label{table:bench_compare}
\end{table}

Many existing KE benchmarks rely on counterfactual constructions, typically involving simple object modifications. While benchmarks like Counterfact \cite{meng2023locatingeditingfactualassociations} and MQuAKE \cite{zhong2024mquakeassessingknowledgeediting} can effectively reflect corrected knowledge, their construction methods often do not align with the true objectives of knowledge editing. In contrast, \dataset{} specifically targets the identification of errors made by LLMs within the e-commerce domain, underscoring the necessity for more practical and nuanced approaches to knowledge editing. This benchmark emphasizes comprehensive editing strategies that preserve the integrity of related knowledge, addressing a significant gap that current benchmarks frequently overlook. By focusing on these critical aspects, future benchmarks can enhance the evaluation of knowledge editing methodologies and ensure that LLMs deliver accurate and up-to-date information across various contexts.

\section{Intrinsic Evaluation}
\begin{enumerate}
    \item \textbf{Wrong Dimension}

\textbf{Product:} \textit{2020 Lenovo V330 15.6" FHD Business Laptop Computer, 8th Gen Intel Quad-Core i7-8550U Up to 4.0GHz, 12GB RAM, 512GB PCIE SSD + 1TB HDD, Windows 10 Professional + EST 500GB External Hard Drive}

\textbf{LLM Output:} \textit{Up to 4GB RAM} 

\textbf{Correct Output:} \textit{Up to 12GB RAM} 

\textbf{Mistake:} Wrong Dimension is particularly prevalent in smaller LLMs, which often struggle with numerical data.

    \item \textbf{Wrong Intention}

\textbf{Product:} \textit{JBL Reference 410 Headphone -Black}

\textbf{LLM Output:} \textit{to listen to music and take calls} 

\textbf{Correct Output:} \textit{To enjoy high-quality audio for immersive listening experiences} 

\textbf{Mistake:} the model provides a basic understanding of the product's functionality but fails to capture the essence of what distinguishes these headphones.

    \item \textbf{Generating own Features}

\textbf{Product:} \textit{Vev Vigano Eco Ceramic Nonstick Frying Pan}

\textbf{LLM Output:} \textit{Durable Construction: The pan is made from high-quality stainless steel.} 

\textbf{Correct Output:} \textit{Durable Eco-Friendly Ceramic Coating: The pan features a high-quality eco-friendly ceramic nonstick surface.} 

\textbf{Mistake:} The model generates its own features by inaccurately claiming that the pan is made from high-quality stainless steel, rather than accurately describing its eco-friendly ceramic nonstick coating.

    \item \textbf{Hallucination}

\textbf{Product:} \textit{RED or BLUE or GREEN Silicone Microwave Popcorn Popper Popcorn Maker CUCINA LAURA}

\textbf{LLM Output:} \textit{[ 1 ]  [ 3 ]  [ 5 ]  [ 7 ]  [ 9 ]  [ 11 ]  [ 13 ]  [ 15 ]  [ 17 ]  [ 19 ]  [ 21 ]  [ 23 ]  [ 25 ]  [ 27 ]} 

\textbf{Correct Output:} \textit{Collapsible Silicone Design for Easy Storage and Microwave Use} 

\textbf{Mistake:} When the model does not have sufficient knowledge about the product, it generates irrelevant and meaningless outputs which is quite common in small-size LLM.
   
\end{enumerate}

\section{KE Methods}
\begin{itemize}
    \item \textbf{FT:} Fine-Tuning (FT) involves adjusting the weights of a pre-trained language model on a specific dataset to improve its performance on targeted tasks.
    \item \textbf{LoRA~\cite{DBLP:conf/iclr/HuSWALWWC22}:} Low-Rank Adaptation (LoRA) modifies model parameters by adding low-rank matrices, enabling efficient fine-tuning with fewer trainable parameters.
    \item \textbf{ROME:} Rank-One Model Editing (ROME) is a technique that allows for the targeted update of specific factual associations within a model by modifying feed-forward weights.
    \item \textbf{MEMIT:} Mass-Editing Memory in a Transformer (MEMIT) expands the capability of language models to update multiple memories simultaneously, allowing for the replacement of outdated information or the addition of new knowledge.
\end{itemize}

\end{document}